\documentclass[conference]{IEEEtran}
\IEEEoverridecommandlockouts
\IEEEpubid{\makebox[\columnwidth]{978-1-7281-0858-2/19/\$31.00~\copyright2019 IEEE \hfill} \hspace{\columnsep}\makebox[\columnwidth]{ }}

\usepackage{cite}
\usepackage{amsmath,amssymb,amsfonts}
\usepackage{algorithmicx, algpseudocode}
\usepackage{graphicx}
\usepackage{textcomp}
\usepackage{xcolor}

\usepackage{amsthm}
\usepackage{bm}
\usepackage{booktabs}
\usepackage{multirow}
\usepackage{subcaption}
\usepackage{thmtools,thm-restate}
\usepackage{url}

\newtheorem{theorem}{Theorem}
\newtheorem{corollary}{Corollary}

\newtheorem{definition}{Definition}

\theoremstyle{remark}
\newtheorem{remark}{Remark}

\begin{document}

\title{Federated Learning with Bayesian Differential Privacy}

\author{\IEEEauthorblockN{Aleksei Triastcyn}
\IEEEauthorblockA{\textit{Artificial Intelligence Lab} \\
\textit{EPFL}\\
Lausanne, Switzerland \\
aleksei.triastcyn@epfl.ch}
\and
\IEEEauthorblockN{Boi Faltings}
\IEEEauthorblockA{\textit{Artificial Intelligence Lab} \\
\textit{EPFL}\\
Lausanne, Switzerland \\
boi.faltings@epfl.ch}
}

\maketitle
\IEEEpubidadjcol

\begin{abstract}
We consider the problem of reinforcing federated learning with formal privacy guarantees. We propose to employ Bayesian differential privacy, a relaxation of differential privacy for similarly distributed data, to provide sharper privacy loss bounds. We adapt the Bayesian privacy accounting method to the federated setting and suggest multiple improvements for more efficient privacy budgeting at different levels. Our experiments show significant advantage over the state-of-the-art differential privacy bounds for federated learning on image classification tasks, including a medical application, bringing the privacy budget below $\varepsilon=1$ at the client level, and below $\varepsilon=0.1$ at the instance level. Lower amounts of noise also benefit the model accuracy and reduce the number of communication rounds.
\end{abstract}


\section{Introduction}
\label{sec:introduction}
The rise of data analytics and machine learning (ML) presents countless opportunities for companies, governments and individuals to benefit from the accumulated data. At the same time, their ability to capture fine levels of detail potentially compromises privacy of data providers. Recent research~\cite{fredrikson2015model, shokri2017membership} suggests that even in a black-box setting it is possible to argue about the presence of individual records in the training set or recover certain features of these records.

To tackle this problem a number of solutions has been proposed. They vary in how privacy is achieved and to what extent data is protected. One approach that assumes privacy at its core is \emph{federated learning} (FL)~\cite{mcmahan2016communication}. In the FL setting, a central entity (\emph{server}) trains a model on user data without actually copying data from user devices. Instead, users (\emph{clients}) update models locally, and the \emph{server} aggregates these updates. 

In spite of all the advantages, federated learning does not provide theoretical privacy guarantees, like it is done by \emph{differential privacy} (DP)~\cite{dwork2006}, which is viewed by many researchers as the privacy gold standard. Initially, DP algorithms focused on sanitising simple statistics, such as mean, median, etc., using a technique known as output perturbation. In recent years, the field made a lot of progress towards the goal of privacy-preserving machine learning, through works on objective perturbation~\cite{chaudhuri2011differentially}, stochastic gradient descend with DP updates~\cite{song2013stochastic}, to more complex and practical methods~\cite{abadi2016deep,papernot2016semi,papernot2018scalable,mcmahan2017learning}.

As shown in recent work~\cite{mcmahan2017learning,geyer2017differentially}, the two approaches can be combined to provide joint benefits. However, unless the number of users is exceedingly high (e.g. in the scenario of a large population of mobile users considered in~\cite{mcmahan2017learning}), differentially private federated learning provides only weak guarantees. Contrary to a wide-spread opinion in machine learning community, values of $\varepsilon$ close to $10$ can hardly be seen as reassurance to a user: for certain types of attacks, an adversary can theoretically reach accuracy of $99.99\%$.

We propose to augment federated learning with a natural relaxation of differential privacy, called \emph{Bayesian differential privacy} (BDP)~\cite{triastcyn2019bayesian}, that provides tighter, and thus, more meaningful guarantees. The main idea of this relaxation is based on the observation that machine learning tasks are often restricted to a particular type of data (for example, finding a film review in the MRI dataset is very unlikely). Moreover, this information, and potentially even some prior distribution of data, is often available to the attacker. While the traditional DP treats all data as equally likely and hides differences by large amounts of noise, BDP calibrates noise to the data distribution. Hence, for any two datasets drawn from the same (arbitrary) distribution, and given the same privacy mechanism with the same amount of noise, BDP provides tighter guarantees than DP. Note that the full knowledge of this distribution is not required, as the necessary statistics can be estimated from data.

We introduce the notion of Bayesian differential privacy in Section~\ref{sec:bayes_dp} and extend it to the federated learning setting in Section~\ref{sec:fed_bayes_dp}. Our experiments (see Section~\ref{sec:evaluation}) show significant advantage, both in privacy guarantees and the model quality.

The main contributions of this paper are the following:
\begin{itemize}
\item we adapt the notion of Bayesian differential privacy to federated learning, including more natural non-i.i.d. settings (Section~\ref{sec:client}), to provide strong theoretical privacy guarantees under minor and practical assumptions;
\item we propose a novel joint accounting method for estimating client-level and instance-level privacy simultaneously and securely (Section~\ref{sec:joint});
\item we experimentally demonstrate advantages of our method, such as shrinking the privacy budget to a fraction of the previous state-of-the-art, and improving the accuracy of the trained models by up to $10\%$ (Section~\ref{sec:evaluation}).
\end{itemize}

\section{Related Work}
\label{sec:related_work}
As machine learning applications become more and more common, various vulnerabilities and attacks on ML models get discovered, based on both passive (for example, model inversion~\cite{fredrikson2015model} and membership inference~\cite{shokri2017membership}) and active adversaries (e.g.~\cite{hitaj2017deep}), raising the need for developing matching defences.

Differential privacy~\cite{dwork2006,dwork2006calibrating} is one of the strongest privacy standards that can be employed to protect ML models from these and other attacks. Since pure $\varepsilon$-DP is hard to achieve in many realistic learning settings, a notion of approximate $(\varepsilon, \delta)$-DP is used across-the-board in machine learning. It is often achieved as a result of applying the Gaussian noise mechanism~\cite{dwork2014algorithmic}.

For a long time, however, even approximate DP remained unachievable in more popular deep learning scenarios. Some earlier attempts~\cite{shokri2015privacy} led to prohibitively high bounds on $\varepsilon$~\cite{abadi2016deep,papernot2016semi} that were later shown to be ineffective against attacks~\cite{hitaj2017deep}. A major step in the direction of bringing privacy loss values down to more practical magnitudes was done by \cite{abadi2016deep} with the introduction of the \emph{moments accountant}, currently a state-of-the-art method for keeping track of the privacy loss during training. Followed by improvements in differentially private training techniques~\cite{papernot2016semi,papernot2018scalable}, it allowed to achieve single-digit DP guarantees ($\varepsilon < 10$) for classic supervised learning benchmarks, such as MNIST, SVHN, and CIFAR.

On the other end of spectrum, McMahan et al.~\cite{mcmahan2016communication} proposed federated learning as one possible solution to privacy issues (among other problems, such as scalability and communication costs). In this setting, privacy is enforced by keeping data on user devices and only submitting model updates to the server. Two of the popular approaches are the federated stochastic gradient descent (\texttt{FedSGD}) and federated averaging (\texttt{FedAvg})~\cite{mcmahan2016communication}, where \emph{clients} do local on-device gradient descent using their data, then send these updates to the \emph{server}, which applies an average update to the model. Privacy can further be enhanced by using secure multi-party computation (MPC)~\cite{yao1982protocols} to allow the server access only average updates of a big group of users and not individual ones. However, MPC or homomorphic encryption do not guarantee robustness against model inversion or membership inference~\cite{fredrikson2015model,shokri2017membership}, because these attacks operate on the resulting model which remains the same. Alternatively, by using differential privacy and the moments accountant, \cite{mcmahan2017learning} and~\cite{geyer2017differentially} attained theoretical client-level privacy guarantees for federated learning settings.

Since federated learning on its own lacks theoretical privacy guarantees, combining it with a more formal notion of privacy is an attractive direction of research. On the other hand, for more complicated deep learning models, differential privacy leads to a poor privacy-utility trade-off, suggesting that pairing federated learning with an alternative notion might prove beneficial. 

Apart from differential privacy, a number of alternative definitions have been proposed over the recent years, such as computational DP~\cite{mironov2009computational}, mutual-information privacy~\cite{mir2012information,wang2016relation}, different versions of concentrated DP (CDP~\cite{dwork2016concentrated}, zCDP~\cite{bun2016concentrated}, tCDP~\cite{bun2018composable}), and R\'enyi DP (RDP)~\cite{mironov2017renyi}. Some other relaxations~\cite{abowd2013differential,schneider2015new,charest2017meaning} tip the balance even further in favour of applicability at the cost of weaker guarantees, for example considering the average-case instead of the worst-case~\cite{triastcyn2019generating}.

In general, important aspects of a privacy notion are composability, accountability, and interpretability. Apart from sharp bounds, the moments accountant is attractive because it operates within the classic notion of $(\varepsilon, \delta)$-DP. Some of the alternative notions of DP also provide tight composition theorems, along with some other advantages, but to the best of our knowledge, they are not broadly used in practice compared to traditional DP (although there are some examples~\cite{geumlek2017renyi}). One of the possible reasons for that is interpretability: parameters of $(\alpha, \varepsilon)$-RDP or $(\mu, \tau)$-CDP are hard to interpret. While it may be difficult to quantify the actual guarantee provided by specific values of $\varepsilon$, $\delta$ of the traditional DP, it is still advantageous that they have a clearer probabilistic interpretation.

In this work, we rely on another relaxation, called Bayesian differential privacy~\cite{triastcyn2019bayesian}. This notion boosts privacy accounting efficiency by utilising the fact that data come from a particular distribution, and not all data are equally likely. At the same time, it maintains the probabilistic interpretation of its parameters $\varepsilon$ and $\delta$. It is worth noting, that unlike some of the relaxations mentioned above, the notion of Bayesian DP provides the worst-case guarantee (under specified conditions) and is not limited to a particular dataset, but rather a particular type of data (e.g. emails, MRI images, etc.), or a mixture of such types, which is a much more permitting assumption.

\section{Preliminaries}
\label{sec:preliminaries}
In this section, we provide necessary definitions, background and notation used in the paper. We also describe the general setting of the problem.

\subsection{Definitions and notation}
\label{sec:definitions}

We use $D, D'$ to represent neighbouring (adjacent) datasets. If not specified, it is assumed that these datasets differ in a single example. Individual examples in a dataset are denoted by $x$ or $x_i$, while the example by which two datasets differ---by $x'$. We assume $D' = D \cup \{x'\}$, whenever possible to do so without loss of generality. 

Since this paper mainly deals with adding noise to gradients (w.r.t. model parameters, neural network weights, etc.), we often refer to non-noised gradients as \emph{non-private outcome}, and denote it by $g, g'$. The private learning outcomes are denoted by $w$. Whenever it is ambiguous, we denote expectation over data (or equivalently, gradients) as $\mathbb{E}_x$, and over the learning outcomes as $\mathbb{E}_w$. Finally, for federated learning scenarios, $u_i$ indicates an update of a user $i$, while $\mathbb{U}$---a set of all user updates.

\begin{definition}
\label{def:differential_privacy}
A randomised function (algorithm) $\mathcal{A}: \mathcal{D} \rightarrow \mathcal{R}$ with domain $\mathcal{D}$ and range $\mathcal{R}$ satisfies $(\varepsilon, \delta)$-differential privacy if for any two adjacent inputs $D, D' \in \mathcal{D}$ and for any set of outcomes $\mathcal{S} \subset \mathcal{R}$ the following holds:
\begin{align*}
	\Pr\left[\mathcal{A}(D) \in \mathcal{S} \right] \leq e^\varepsilon \Pr\left[\mathcal{A}(D') \in \mathcal{S} \right] + \delta.
\end{align*}
\end{definition}

\begin{definition}
\label{def:privacy_loss}
The privacy loss $L_{\mathcal{A}}$ of a randomised algorithm $\mathcal{A}: \mathcal{D} \rightarrow \mathcal{R}$ for outcome $w \in \mathcal{R}$ and datasets $D, D' \in \mathcal{D}$ is given by:
\begin{align*}
	L_{\mathcal{A}}(w, D, D') = \log\frac{\Pr\left[\mathcal{A}(D) = w \right]}{\Pr\left[\mathcal{A}(D') = w \right]}.
\end{align*}
\end{definition}

For notational simplicity, we omit the designation $\mathcal{A}$, i.e. we use $L(w, D, D')$ (or simply $L$) for the privacy loss random variable, and $p(w | D)$, and $p(w | D')$ for the outcome probability distributions for given datasets. Also, note that the privacy loss random variable $L$ is distributed by drawing $w \sim p(w | D)$ (see~\cite[Section 2.1 and Definition 3.1]{dwork2016concentrated}), which helps linking it to well-known divergences.

\begin{definition}
\label{def:gaussian_mechanism}
The Gaussian noise mechanism achieving $(\varepsilon, \delta)$-DP, for a function $f: \mathcal{D} \rightarrow \mathbb{R}^m$, is defined as
\begin{align*}
	\mathcal{M}(D) = f(D) + \mathcal{N}(0, I \sigma^2),
\end{align*}
where $\sigma > C \sqrt{2\log\frac{1.25}{\delta}} / \varepsilon$ and $C = \max_{D, D'} \|f(D) - f(D')\|$ is the L2-sensitivity of $f$.
\end{definition}
For more details on differential privacy and the Gaussian mechanism, we refer the reader to~\cite{dwork2014algorithmic}.

We will also need the definition of R\'enyi divergence:
\begin{definition}
\label{def:renyi_divergence}
R\'enyi divergence of order $\lambda$ between distributions $P$ and $Q$, denoted as $D_\lambda (P \| Q)$ is defined as
\begin{align*}
	D_\lambda (P \| Q) &= \frac{1}{\lambda - 1} \log \int p(x) \left[\frac{p(x)}{q(x)}\right]^{\lambda-1} dx \nonumber \\
		&= \frac{1}{\lambda - 1} \log \int q(x) \left[\frac{p(x)}{q(x)}\right]^{\lambda} dx,
\end{align*}
where $p(x)$ and $q(x)$ are corresponding density functions of $P$ and $Q$.
\end{definition}
Analytic expressions for R\'enyi divergence exist for many common distributions and can be found in~\cite{gil2013renyi}. Van Erven and Harremos~\cite{van2014renyi} provide a good survey of R\'enyi divergence properties in general.

\subsection{Setting}
\label{sec:setting}
In the first part of the paper, while describing the concept of Bayesian differential privacy, we consider a general iterative learning algorithm, such that each iteration $t$ produces a non-private learning outcome $x^{(t)}$ (e.g. a gradient over a batch of data). In the second half of the paper, we consider the equivalent federated learning setting, where each communication round $t$ produces a set of non-private learning outcomes $u_i^{(t)}$, one for each client $i$. 

The non-private outcome, in both scenarios, gets transformed into a private learning outcome $w^{(t)}$ that is used as a starting point for the next iteration or communication round. The learning outcome can be made private by different means, but in this work we consider the most common approach of applying an additive noise mechanism (e.g. a Gaussian noise mechanism). We denote the distribution of private outcomes by $p(w^{(t)} | w^{(t-1)}, D)$ (we assume the Markov property of the learning process for brevity of notation, but it is not necessary in general) or $p(w^{(t)} | w^{(t-1)}, \mathbb{U})$, depending on the scenario. 

The process can run on subsamples of data or subsets of clients, in which case $w^{(t)}$ comes from the distribution $p(w^{(t)} | w^{(t-1)}, B^{(t)})$, where $B^{(t)}$ is a batch of data used for parameters update in iteration $t$, or $p(w^{(t)} | w^{(t-1)}, \mathbb{U}^{(t)})$, where $\mathbb{U}^{(t)}$ is a set of updates from users participating in the communication round $t$. In these cases, privacy is amplified through sampling~\cite{balle2018privacy}. 

For each iteration, we would like to compute a quantity $c_t$ (we call it a \emph{privacy cost}) that accumulates over the learning process and allows to compute privacy loss bounds $\varepsilon, \delta$ using concentration inequalities. The overall privacy accounting workflow does not significantly differ from prior work, but is in fact a generalisation of the well-known moments accountant~\cite{abadi2016deep}. Importantly, it is not tied to a specific learning algorithm or a class of algorithms, as long as one can map it to the above setting.

\subsection{Motivation}
\label{sec:motivation}
Before we proceed, we find it important to motivate the research and usage of alternative definitions of privacy. The primary reason for this is that the complexity of the concept of differential privacy often leads to misunderstanding or overestimation of the guarantees it provides. And while we do not fully tackle the problem of interpretability, we provide a simple example below that allows to better judge the quality of provided guarantees.

Consider the state-of-the-art differentially private machine learning models~\cite{abadi2016deep,papernot2018scalable}. In order to come close to the non-private accuracy (say within $10\%$ of it), all of the reported models stretch their privacy budget to $\varepsilon > 2$ (for a reasonably low $\delta$), while in many cases it goes up to $\varepsilon > 5$. In real-world applications, it can even be larger than $10$\footnote{\url{https://www.wired.com/story/apple-differential-privacy-shortcomings/}}. These numbers seem small, and thus, may often be overlooked. But let us present an alternative interpretation.

What we are interested in is the change in the posterior distribution of the attacker after they see the private model compared to prior~\cite{mironov2017renyi,bun2017teaser}. Let us consider the stronger, pure DP for simplicity. According to the definition of $\varepsilon$-DP:
\begin{align*}
	\frac{p(D | w)}{p(D' | w)} \leq e^\varepsilon \frac{p(D)}{p(D')}.
\end{align*}
Assume the following specific example. The datasets $D, D'$ consist of income values for residents of a small town. There is one individual $x'$ whose income is orders of magnitude higher than the rest, and whose residency in the town is what the attacker wishes to infer. The attacker observes the mean income $w$ sanitised by a differentially private mechanism with $\varepsilon=\varepsilon_0$. If the individual is not present in the dataset, the probability of $w$ being above a certain threshold is extremely small. On the contrary, if $x'$ is present, this probability is higher (say it is equal to $r$). The attacker takes a Bayesian approach, computes the likelihood of the observed value under each of the two assumptions and the corresponding posteriors given a flat prior. The attacker then concludes that the individual is present in the dataset and is a resident.

By the above expression, $r$ can only be $e^{\varepsilon_0}$ times larger than the corresponding probability without $x'$. But if the $r e^{-\varepsilon_0}$ is small enough, then the probability $P(A)$ of the attacker's guess being correct is as high as $\frac{r}{r + re^{-\varepsilon_0}}$ or, equivalently,
\begin{align}
\label{eq:motivation}
	P(A) = \frac{1}{1 + e^{-\varepsilon}}.
\end{align}

To put it in perspective, for a DP algorithm with $\varepsilon = 5$, the upper bound on the accuracy of this attack is as high as $99.33\%$. For $\varepsilon = 8$, it is $99.97\%$. For $\varepsilon = 10$, $99.995\%$. Remember that we used an uninformative flat prior, and for a more informed attacker these numbers could be even larger. 

In a more realistic scenario, even without any privacy protection, this high accuracy is not likely to be achieved by the attacker. So such guarantee is hardly better than no guarantee, and cannot be seen as reassuring. Thus, we want to encourage the discussion and search for more meaningful privacy definitions or DP relaxations for machine learning and federated learning. One such relaxation we present and explore in this paper.

\section{Bayesian Differential Privacy}
\label{sec:bayes_dp}
In this section, we describe \emph{Bayesian differential privacy (BDP)}, accompanied by a practical privacy loss accounting method. We restate just the main results necessary for the following section, while all the details, proofs and experimental evaluation of BDP can be found in~\cite{triastcyn2019bayesian}.

\begin{definition}
\label{def:bayes_dp}
A randomised function (algorithm) $\mathcal{A}: \mathcal{D} \rightarrow \mathcal{R}$ with domain $\mathcal{D}$ and range $\mathcal{R}$ satisfies $(\varepsilon, \delta)$-Bayesian differential privacy if for any two adjacent datasets $D, D' \in \mathcal{D}$, \emph{differing in a single data point $x' \sim p(x)$}, and for any set of outcomes $\mathcal{S} \subset \mathcal{R}$ the following holds:
\begin{align}
	\Pr\left[\mathcal{A}(D) \in \mathcal{S} \right] \leq e^\varepsilon \Pr\left[\mathcal{A}(D') \in \mathcal{S} \right] + \delta,
\end{align}
\end{definition}

To derive tighter sequential composition, we use an alternative definition that implies the above:
\begin{align}
	\Pr[L(w, D, D') \geq \varepsilon] \leq \delta,
\end{align}
where probability is taken over the randomness of the outcome $w$ and the additional example $x'$.

This definition is very close to the original definition of DP, except that it also takes into account the randomness of $x'$. Hence, the basic properties are similar to those of DP, although BDP does not provide guarantees in all the scenarios where DP does (e.g. when the distribution $p(x)$ is non-stationary, it is possible that BDP underestimates the actual privacy loss).

While Definition~\ref{def:bayes_dp} does not specify the distribution of any point in the dataset other than the additional example $x'$, it is natural and convenient to assume that all examples in the dataset are drawn from the same distribution $p(x)$. This holds in many real-world applications, including all applications evaluated in this paper, and it allows using sampling techniques instead of requiring knowing the true distribution.

We also assume that all data points are exchangeable~\cite{aldous1985exchangeability}, i.e. any permutation of data points has the same joint probability. It enables tighter accounting for iterative mechanisms, and is naturally satisfied in the considered scenarios.

Since basic composition is not enough to provide tight privacy bounds for iterative federated learning algorithms in which we are interested, let us present two theorems, generalising upon the moments accountant routine. We use it as a foundation for the privacy accounting framework for federated learning presented in the next section.

\begin{restatable}[Advanced Composition]{theorem}{composition}
\label{thm:advanced_composition}
Let a learning algorithm run for $T$ iterations. Denote by $w^{(1)} \ldots w^{(T)}$ a sequence of private learning outcomes obtained at iterations $1,\ldots,T$, and $L^{(1:T)}$ the corresponding total privacy loss. Then,
\begin{align*}
	\mathbb{E}\left[e^{\lambda L^{(1:T)}}\right] = \prod_{t=1}^T \mathbb{E}_x \left[ e^{\lambda D_{\lambda+1} (p_t \| q_t)} \right],
\end{align*}
where $p_t = p(w^{(t)} | w^{(t-1)}, D)$, $q_t = p(w^{(t)} | w^{(t-1)}, D')$, and $D_{\lambda+1} (p_t \| q_t)$ is R\'enyi divergence between $p_t$ and $q_t$.
\end{restatable}

We denote the logarithm of the quantity inside the product in Theorem~\ref{thm:advanced_composition} as $c_t(\lambda)$ and call it the \emph{privacy cost} of the iteration, or communication round, $t$:
\begin{align}
\label{eq:privacy_cost}
	c_t(\lambda) = \log \mathbb{E} \left[ e^{\lambda D_{\lambda+1} (p_t \| q_t)} \right]
\end{align}

The privacy cost of the whole learning process is then a sum of the costs of each iteration.
\begin{theorem}
\label{thm:eps_delta_relation}
Let the algorithm produce a sequence of private learning outcomes $w^{(1)} \ldots w^{(T)}$ using a known probability distribution $p(w^{(t)} | w^{(t-1)}, D)$. Then, $\varepsilon$ and $\delta$ are related as
\begin{align*}
	\log \delta \leq \sum_{t=1}^T c_t(\lambda) - \lambda \varepsilon.
\end{align*}
\end{theorem}

\begin{corollary}
\label{thm:eps_from_delta}
Under the conditions above, for a fixed $\delta$:
\begin{align*}
	\varepsilon \leq \frac{1}{\lambda} \sum_{t=1}^T c_t(\lambda) - \frac{1}{\lambda} \log \delta.
\end{align*}
\end{corollary}

Theorem~\ref{thm:eps_delta_relation} provides an efficient privacy accounting algorithm. During training, we compute the privacy cost $c_t(\lambda)$ for each iteration $t$, accumulate it, and then use to compute $\varepsilon, \delta$ pair. This process is ideologically close to that of the moment accountant, but accumulates a different quantity (note the change from the privacy loss random variable to R\'enyi divergence and expectation over data).

Computing $c_t(\lambda)$ precisely requires access to the prior distribution of data $p(x)$, which is unrealistic. Therefore, we need an estimator for $\mathbb{E}[e^{\lambda D_{\lambda+1} (p_t \| q_t)}]$. Moreover, since Chernoff bound, which our Theorem~\ref{thm:eps_delta_relation} is based on, only holds for the true expectation value, we have to take into account the estimator error. To solve this, we employ a Bayesian view of the estimation problem~\cite{oliphant2006bayesian} and use the upper confidence bound of the expectation estimator.

Let us define the following $m$-sample estimator of $c_t(\lambda)$:
\begin{align}
\label{eq:estimator}
	\hat{c}_t(\lambda) = \log \left[ M(t) + \frac{F^{-1}(1 - \delta', m - 1)}{\sqrt{m - 1}} S(t) \right],
\end{align}
where $M(t)$ and $S(t)$ are the sample mean and the sample standard deviation of $e^{\lambda \hat{D}_{\lambda+1}^{(t)} }$, $F^{-1}(1-\delta', m - 1)$ is the inverse of the Student's $t$-distribution CDF at $1-\delta'$ with $m - 1$ degrees of freedom, and
\begin{align*}
	& \hat{D}_{\lambda+1}^{(t)}  = \max \left\{ D_{\lambda+1} (\hat{p}_t \| \hat{q}_t),~D_{\lambda+1} (\hat{q}_t \| \hat{p}_t) \right\}, \\
	& \hat{p}_t = p(w^{(t)}~|~w^{(t-1)}, B^{(t)}), \\
	& \hat{q}_t = p(w^{(t)}~|~w^{(t-1)}, B^{(t)} \setminus \{x_i\}).
\end{align*}
One can show that, for continuous distributions $\hat{p}_t$ and $\hat{q}_t$, $\hat{c}_t(\lambda)$ overestimates the true privacy cost $c_t(\lambda)$ with probability $1-\delta'$. Therefore, the probability of underestimation $\delta'$ can be fixed upfront and incorporated in $\delta$.

\begin{remark}
This step changes the interpretation of $\delta$ in Bayesian differential privacy compared to the traditional DP. Apart from the probability of the privacy loss exceeding $\varepsilon$, e.g. in the tails of its distribution, it also incorporates our uncertainty about the true data distribution (in other words, the probability of underestimating the true expectation because of not observing enough data samples). It can be intuitively understood as accounting for unlikely or unobserved data in $\delta$, rather than in $\varepsilon$ by adding more noise.
\end{remark}

\begin{remark}
For discrete outcomes, a different estimator needs to be derived to meet the same bound. The process of obtaining it is identical to the one above, with the only change in the maximum entropy distribution.
\end{remark}

\begin{remark}
There are other differences compared to the classic DP, such as allowance for unbounded sensitivity or estimator privacy, that are not discussed in this work. More details can be found in~\cite{triastcyn2019bayesian}.
\end{remark}

\textbf{Gaussian Noise Mechanism.} Consider the subsampled Gaussian noise mechanism~\cite{dwork2014algorithmic}. The outcome distribution $p(w^{(t)}~|~w^{(t-1)}, D)$ in this case is the mixture of two Gaussians $(1-q) \mathcal{N}(g_t, \sigma^2) + q \mathcal{N}(g_t', \sigma^2)$, where $g_t$ and $g_t'$ are non-private outcomes at the iteration $t$ (e.g. gradients), $\sigma$ is the noise parameter, and $q$ is the data sampling probability. Plugging the outcome distribution into the formula for R\'enyi divergence, we get the following result for the privacy cost.
\begin{restatable}{theorem}{gaussian}
\label{thm:gauss_privacy_cost}
Given the Gaussian noise mechanism with the noise parameter $\sigma$ and subsampling probability $q$, the privacy cost for $\lambda \in \mathbb{N}$ at iteration $t$ can be expressed as
\begin{align*}
	c_t(\lambda) = \max\{c_t^{L}(\lambda), c_t^{R}(\lambda)\},
\end{align*}
where
\begin{align*}
	&c_t^{L}(\lambda) = \log \mathbb{E}_x \left[ \mathbb{E}_{k \sim B(\lambda+1, q)} \left[e^{\frac{k^2 - k}{2\sigma^2} \|g_t - g_t'\|^2} \right] \right], \\
	&c_t^{R}(\lambda) = \log \mathbb{E}_x \left[ \mathbb{E}_{k \sim B(\lambda, q)} \left[e^{\frac{k^2 + k}{2\sigma^2} \|g_t - g_t'|^2} \right] \right],
\end{align*}
and $B(\lambda, q)$ is the binomial distribution with $\lambda$ experiments and the probability of success $q$. 
\end{restatable}

\section{Federated Learning with Bayesian Differential Privacy}
\label{sec:fed_bayes_dp}
In this section, we adapt the Bayesian differential privacy framework and its accounting method to guarantee the client-level privacy, the level most frequently addressed in the literature. We then justify and explore the instance-level privacy and two different techniques for accounting it. Finally, we propose a method to jointly account instance-level and client-level privacy for the \texttt{FedSGD} algorithm in order to provide the strongest trade-off between utility and privacy guarantees.

\subsection{Client privacy}
\label{sec:client}
When it comes to reinforcing federated learning with differential privacy, the foremost attention is given to the client-level privacy~\cite{mcmahan2017learning,geyer2017differentially}. The goal is to hide the presence of a single user, or to be more specific, to bound the influence of any single user on the learning outcome distribution (i.e. the distribution of the model parameters).

Under the classic DP~\cite{mcmahan2017learning,geyer2017differentially}, the privacy is enforced by clipping all user updates $u_i$ to a fixed $L2$-norm threshold $C$ and then adding Gaussian noise with the variance $C^2 \sigma^2$. The noise parameter $\sigma$ is calibrated to bound the privacy loss in each communication round, and then the privacy loss is accumulated across the rounds using the moments accountant~\cite{abadi2016deep}.

We use the same privacy mechanism, but employ the Bayesian accounting method instead of the moments accountant. Intuitively, our accounting method should have a significant advantage over the moments accountant in the settings where data is distributed similarly across the users because in this case their updates would be in a strong agreement. In order to map the Bayesian differential privacy framework to this setting, let us introduce some notation.

Let $N$ denote the number of clients in the federated learning system. Every client $i$ computes and sends to the server a model update $u_i \sim p_i(u)$ drawn from the client's update distribution $p_i(u)$. Considering individual client distributions ensures that our approach is applicable to non-i.i.d. settings that are natural in the federated learning context. Generally, not all users participate in a given communication round. We denote the probability of a user $i$ participating in the round by $\alpha_i$. Thus, the overall update distribution is given by a mixture:
\begin{align}
	p(u) = \sum_{i=1}^{N} \alpha_i p_i(u).
\end{align}
In our experiments, we fix $\alpha_1 = \alpha_2 = \ldots = \alpha_N = \alpha$.

To match the notation above, let $w_t$ indicate the privacy-preserving model update:
\begin{align}
	w_t = \mathcal{A}(\{u_i | u_i \in \mathbb{U}^{(t)} \}),
\end{align}
where $\mathcal{A}(\{u_i | u_i \in \mathbb{U}^{(t)} \}) = \frac{1}{|\mathbb{U}^{(t)}|} \sum_i u_i + \mathcal{N}(0, C^2\sigma^2)$ in the case of Gaussian mechanism, and $\mathbb{U}^{(t)}$ is the set of updates from users participating in the round $t$.

To bound $\varepsilon$ and $\delta$ of Bayesian differential privacy, one needs to compute $c_t(\lambda) = \max\{c_t^{L}(\lambda), c_t^{R}(\lambda)\}$, where
\begin{align*}
	&c_t^{L}(\lambda) = \log \mathbb{E}_u \left[ e^{\lambda D_{\lambda+1} (p_t \| q_t)} \right], \\
	&c_t^{R}(\lambda) = \log \mathbb{E}_u \left[ e^{\lambda D_{\lambda+1} (q_t \| p_t)} \right],
\end{align*}
and
\begin{align*}
	&p_t = p(w^{(t)} | w^{(t-1)}, \mathbb{U}^{(t)}) \\
	&q_t = p(w^{(t)} | w^{(t-1)}, \mathbb{U}^{(t)} \setminus \{u\})
\end{align*}
Since the randomness of $w$ comes from the subsampled Gaussian noise mechanism, we use Theorem~\ref{thm:gauss_privacy_cost}, in combination with user sampling and the estimator~\eqref{eq:estimator} for both expressions, to obtain $\hat{c}_t(\lambda)$ that upper-bounds $c_t(\lambda)$ with high probability.

Finally, we use Theorems~\ref{thm:advanced_composition} and~\ref{thm:eps_delta_relation} to compute $\varepsilon$ and $\delta$. The required assumption of exchangeability is naturally satisfied because users are sampled independently and uniformly.

\subsection{Instance privacy}
\label{sec:instance}
As noted above, the same privacy mechanism can be used in conjunction with the moments accountant to get the classic DP guarantees~\cite{mcmahan2017learning,geyer2017differentially}. In this case, $(\varepsilon, \delta)$-DP at the client level implies the same guarantee at the instance level (i.e. bounding the influence of a single data point). However, it does not hold for BDP. Moreover, the same privacy guarantee may not be meaningful at the instance level. For example, $\delta=10^{-3}$ might be reasonable for $100$ clients, but if a client has tens of thousands of data points, it is not a reasonable failure probability at the data point level.

At the same time, instance privacy is extremely important in some scenarios. Imagine federated training on medical data from different hospitals: while a hospital participation may be public knowledge, individual patients data must be protected at the highest degree. Another reason for considering instance-level privacy is that it provides an additional layer of protection for users in case of an untrusted curator.

In order to get tighter instance privacy guarantees, we apply the subsampled Gaussian noise mechanism to gradient computation on user devices. The accounting follows the same procedure as described above, except that the noise parameter $\sigma$ and the sampling probability $q$ may be different, depending on which of the settings described below is used.

There are two possible accounting schemes:
\subsubsection{Sequential accounting}
\label{sec:sequential}
Part of the accounting is performed locally on user devices and part on the server. Overall privacy cost is equivalent to the centralised training with the data sampling probability $q=\frac{B_i}{N}$, where $N$ is the total number of data points across all users, and $B_i$ is the local batch size.

The process proceeds as follows. At each communication round, the server sends $N$ to participating clients, every client performs private gradient updates, computes $\hat{c}_t(\lambda)$, and sends it to the server. The server then aggregates the sum of $\hat{c}_t(\lambda)$ from all users. Since the privacy costs are data-dependent, it is possible to use secure multi-party computation to allow the server know the sum without learning individual costs.

The disadvantage of this method is that every participating client learns the total number of data points, and especially in the settings with a small number of users it may not be desirable. Furthermore, the obtained bounds apply to the commonly learnt model but not to the individual updates of each user, requiring them to maintain a separate local bound. These issues are addressed by parallel accounting.

\subsubsection{Parallel accounting}
\label{sec:parallel}
In this scheme, every client computes $\hat{c}_t(\lambda)$ using $q=\frac{B_i}{N_i}$, where $N_i$ is the local dataset size of the client. Consequently, since $N_i \leq N$, the privacy costs will be higher. But this is compensated by using parallel composition instead of sequential: the server aggregates the maximum of $\hat{c}_t(\lambda)$ over all users. Again, using secure multi-party computation is possible to prevent the server from learning individual privacy costs.

Parallel composition is applicable in this scenario because user updates within a single round are independent. However, the server needs to sum up maximum privacy costs over the rounds because updates are dependent on previous rounds.

As we show in Section~\ref{sec:evaluation}, parallel accounting may require more communication rounds to converge to the same quality solution with the same privacy guarantee. The gap is more notable on non-identically distributed data.

\subsection{Joint privacy}
\label{sec:joint}
Instance privacy provides tighter and more meaningful guarantees for every data point contribution to the trained model. Nevertheless, there is a downside: adding noise both during on-device gradient descent as well as during the averaging phase on the server results in slow convergence or complete divergence of the federated learning algorithm.

To tackle this problem, we propose \emph{joint accounting}, where the noise added on the client side is re-counted towards the client-level privacy guarantee.

The main idea of joint accounting is that a client update received by the server is already noisy when instance privacy is enforced, and instead of adding more noise the server can re-count existing noise to compute the client-level bound\footnote{A version of this technique for differential privacy has also been explored in a master's thesis project with Nikolaos Tatarakis.}. The only problem: the server cannot sample non-private client updates to estimate $c_t(\lambda)$ because it no longer has access to their distribution. 

Fortunately, the inner expectation in $c_t^{L}(\lambda)$ and $c_t^{R}(\lambda)$ can be computed locally, suggesting the following procedure. Every client computes $\hat{D}_{\lambda+1}^{(t)}$ (Eq.~\ref{eq:estimator}) with $\hat{p}_t$ and $\hat{q}_t$ being the private outcome distributions with and without their entire update. Then, the server computes $M(t)$, $S(t)$, and $\hat{c}_t(\lambda)$ by simple averaging. Additionally, one can implement this averaging step with secure multi-party computation to further privacy protection. For the moment, however, it can only be used with \texttt{FedSGD}, and not \texttt{FedAvg}, because every noisy step in \texttt{FedAvg} would change the point at which the gradient is computed, potentially leading to a different gradient distribution or underestimated total noise variance.

Using joint accounting allows to achieve tight instance and client privacy guarantees, and at the same time, preserve the speed of convergence almost at the same level as the client-privacy-only solution (see Section~\ref{sec:evaluation_joint}).

\section{Evaluation}
\label{sec:evaluation}
In this section, we provide results of the experimental evaluation of our approach. We begin by describing the datasets we used, as well as the setting details shared by all experiments. The subsequent structure follows that of the previous section. We first evaluate the client-level privacy by comparing accuracy and privacy guarantees of the traditional DP method~\cite{geyer2017differentially} to ours (Section~\ref{sec:evaluation_client}). Then, in Section~\ref{sec:evaluation_instance}, we perform experiments on the two proposed methods of instance privacy accounting. Finally, Section~\ref{sec:evaluation_joint} describes the results of the joint accounting approach.

\subsection{Experimental setup}
\label{sec:evaluation_setup}
We perform experiments on two datasets. The first dataset is MNIST~\cite{lecun1998gradient}. It is a standard image classification task widely used in machine learning research. More specifically, it is a handwritten digit recognition dataset consisting of 60000 training examples and 10000 test examples, each example is a 28x28 greyscale image. The second is the APTOS 2019 Blindness Detection challenge dataset\footnote{\url{https://www.kaggle.com/c/aptos2019-blindness-detection/overview/description}} (in figures, tables and text, we refer to this dataset as \emph{Retina} or \emph{APTOS}). It consists of 3662 retina images taken using fundus photography. The images are labelled by clinicians to reflect the severity of diabetic retinopathy on the scale from 0 to 4. Unlike other datasets commonly evaluated in the privacy literature~\cite{abadi2016deep,mcmahan2017learning,geyer2017differentially}, this one actually has more serious implications of a privacy leak.

All experiments have the following general setup. There is a number of clients (100, 1000, or 10000), each holding a subset of data, and the server that coordinates federated training of the shared model. Some setups with a higher number of users will entail repetition of data, like in~\cite{geyer2017differentially}, which is a natural scenario in some applications, e.g. shared or very similar images on different smartphones. In MNIST experiments, each user holds 600 examples. For the APTOS dataset, we use data augmentation techniques (e.g. random cropping, resizing, etc.) to obtain a larger training set, and then split it such that every client gets $\sim$350 images.

Testing is performed on the official test split for MNIST, and on the first 500 samples in case of APTOS.

While the parameters of the training vary based on experiments, the models remain the same. For MNIST, we use a simple CNN with two convolutional layers and two fully connected layers (similar to the one described in the TensorFlow tutorial\footnote{\url{https://www.tensorflow.org/tutorials/images/deep_cnn}}). In case of APTOS, due to the small dataset size and a harder learning task, we employ ResNet-50~\cite{he2016deep} pre-trained on ImageNet~\cite{deng2009imagenet} and re-train only the last fully-connected layer of the network. We do not do extensive hyper-parameter tuning in general, since we are interested in relative performance of private models compared to non-private ones rather than the best classification accuracy, and thus, our non-private baseline results may not always match the ones reported in~\cite{mcmahan2016communication}. For the same reason, we restrict the number of communication rounds ($\leq 300$) and use $\texttt{FedSGD}$ instead of $\texttt{FedAvg}$, although all the methods except for joint accounting are compatible with $\texttt{FedAvg}$.

One of the important aspects of federated learning is that data might not be distributed identically among users. In agreement with previous work~\cite{mcmahan2016communication,geyer2017differentially}, we include experiments in both \emph{i.i.d.} and \emph{non-i.i.d.} settings for MNIST, because it allows for a natural non-identical split. More specifically, in the i.i.d. setting, every user is assigned a subset of uniformly sampled examples. In the non-i.i.d. setting, we follow the same scheme as~\cite{mcmahan2016communication} and~\cite{geyer2017differentially}: splitting the dataset on shards of 300 points within the same class and then assigning 2 random shards to each client. The scenario of 100 clients with non-identically distributed data is particularly hard for privacy applications: there are ${10 \choose 2} = 45$ possible digit combinations that clients can hold and only 100 clients, meaning that some clients might be easily distinguishable by their data distribution. Therefore, it is important to note that it may not be possible to obtain a reasonable privacy bound in this scenario without seriously compromising accuracy.

The privacy accounting is performed by two methods. To obtain the bounds on $\varepsilon$ and $\delta$ of differential privacy, we use the moments accountant~\cite{abadi2016deep}, the state-of-the-art DP accounting method. In the case of Bayesian differential privacy, we follow the technique described in Sections~\ref{sec:bayes_dp} and~\ref{sec:fed_bayes_dp}: we sample a number of user updates (or gradients for instance privacy), estimate an upper bound on the privacy cost, and use Chernoff inequality to compute the corresponding pair of $\varepsilon, \delta$.

\subsection{Client privacy}
\label{sec:evaluation_client}

\begin{table}
	\caption{Accuracy and privacy guarantees (reported as a pair $(\varepsilon, \delta)$) on MNIST, non-i.i.d. setting.}
	\label{tab:mnist_noniid}
	\centering
	\begin{tabular}{ l | c | c c | c c }
		\toprule
								&  \multicolumn{3}{c|}{Accuracy} 	& \multicolumn{2}{c}{Privacy} 		\\
		\midrule
		{\bf Clients}		& {\bf Baseline}		& {\bf DP} 		& {\bf BDP} 				& {\bf DP} 				& {\bf BDP} 			\\
		\midrule
		100					& $97\%$ 				& $78\%$ 			& $\bm{88\%}$			& $(8, 10^{-3})$ 		 & $\bm{(4.0, 10^{-3})}$ 	\\
		1K						& $98\%$ 			& $95\%$ 		& $\bm{96\%}$			& $(3, 10^{-5})$ 		 & $\bm{(1.5, 10^{-5})}$ 	\\
		10K					& $99\%$ 			& $96\%$ 		& $\bm{97\%}$			& $(1, 10^{-6})$ 		 & $\bm{(0.6, 10^{-6})}$ 	\\
		\bottomrule
	\end{tabular}
\end{table}

\begin{table}
	\caption{Accuracy and privacy guarantees (reported as a pair $(\varepsilon, \delta)$) on MNIST, i.i.d. setting.}
	\label{tab:mnist_iid}
	\centering
	\begin{tabular}{ l | c | c c | c c }
		\toprule
								&  \multicolumn{3}{c|}{Accuracy} 	& \multicolumn{2}{c}{Privacy} 		\\
		\midrule
		{\bf Clients}		& {\bf Baseline}		& {\bf DP} 		& {\bf BDP} 				& {\bf DP} 				& {\bf BDP} 			\\
		\midrule
		100					& $97\%$ 				& $86\%$ 		& $\bm{92\%}$ 		& $(8, 10^{-3})$ 		& $\bm{(2.0, 10^{-3})}$ 	\\
		1K						& $98\%$ 			& $97\%$ 	 		& $97\%$					& $(3, 10^{-5})$ 		& $\bm{(1.0, 10^{-5})}$ 	\\
		10K					& $99\%$ 			& $97\%$ 			& $\bm{98\%}$			& $(1, 10^{-6})$ 	 	& $\bm{(0.5, 10^{-6})}$ 	\\
		\bottomrule
	\end{tabular}
\end{table}

\begin{figure*}
	\centering
	\begin{subfigure}{0.32\textwidth}
		\includegraphics[width=\linewidth]{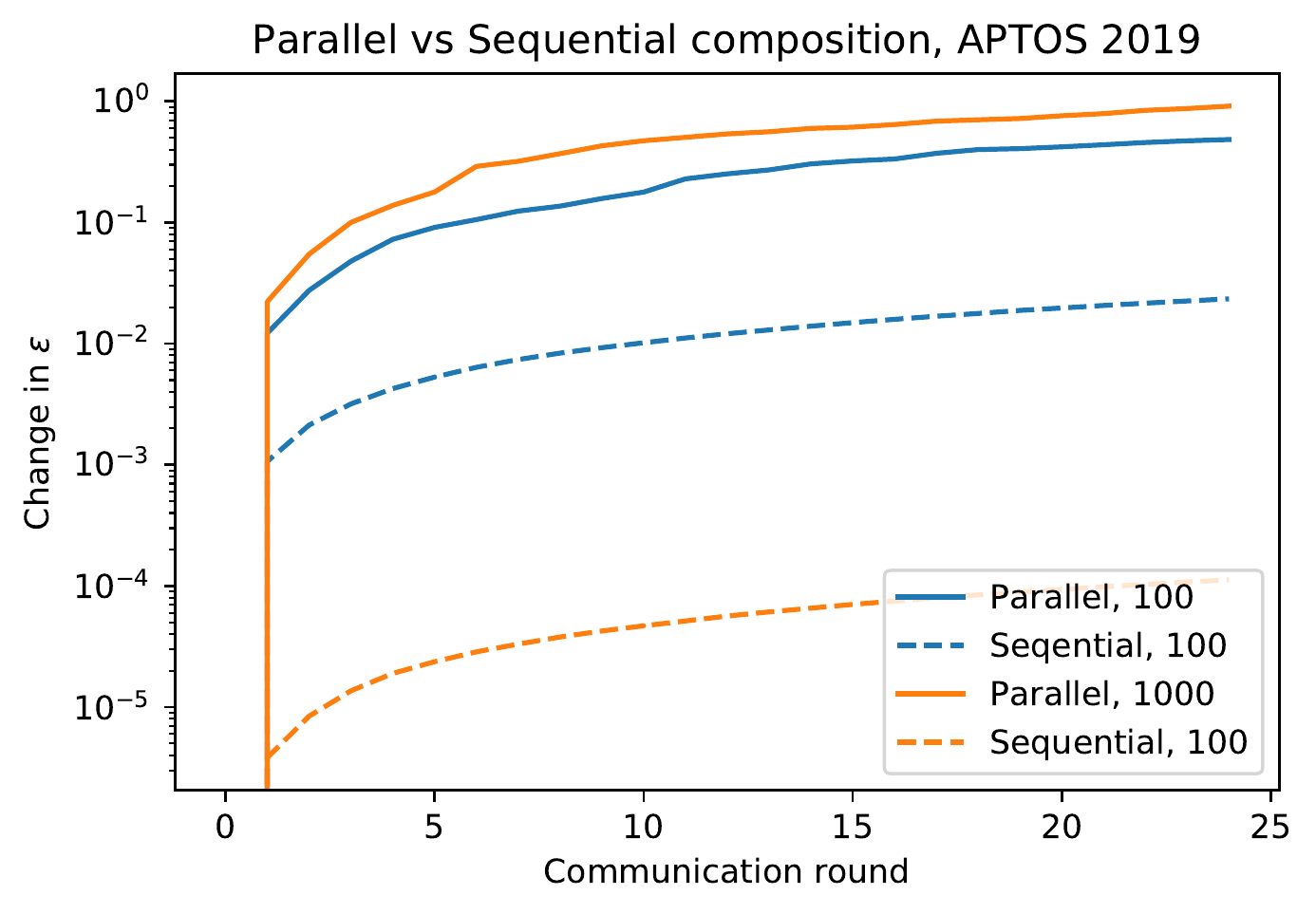}
		\caption{APTOS 2019}
		\label{fig:eps_par_seq_retina}
	\end{subfigure}
	\begin{subfigure}{0.32\textwidth}
		\includegraphics[width=\linewidth]{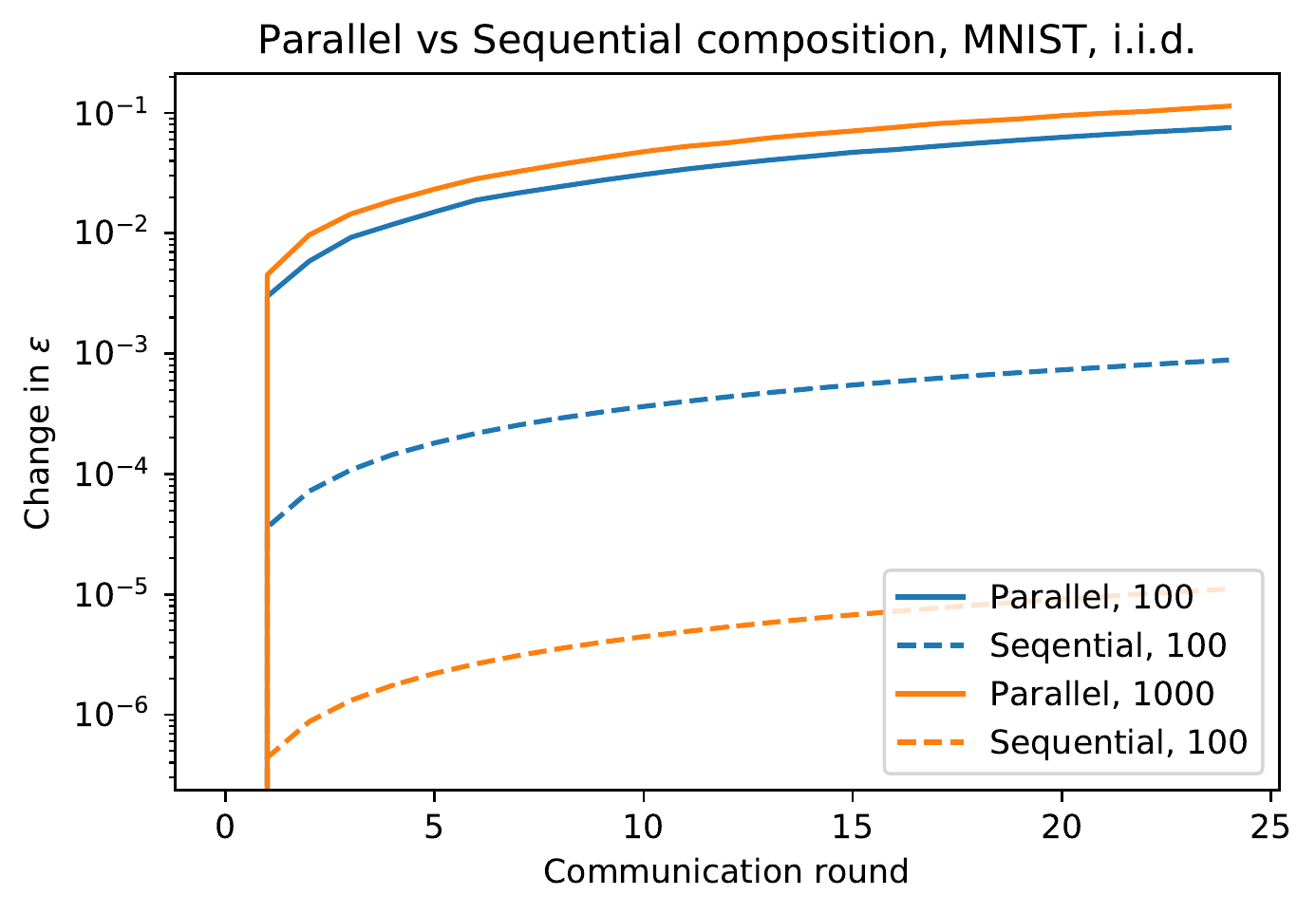}
		\caption{MNIST, i.i.d.}
		\label{fig:eps_par_seq_mnist_iid}
	\end{subfigure}
	\begin{subfigure}{0.32\textwidth}
		\includegraphics[width=\linewidth]{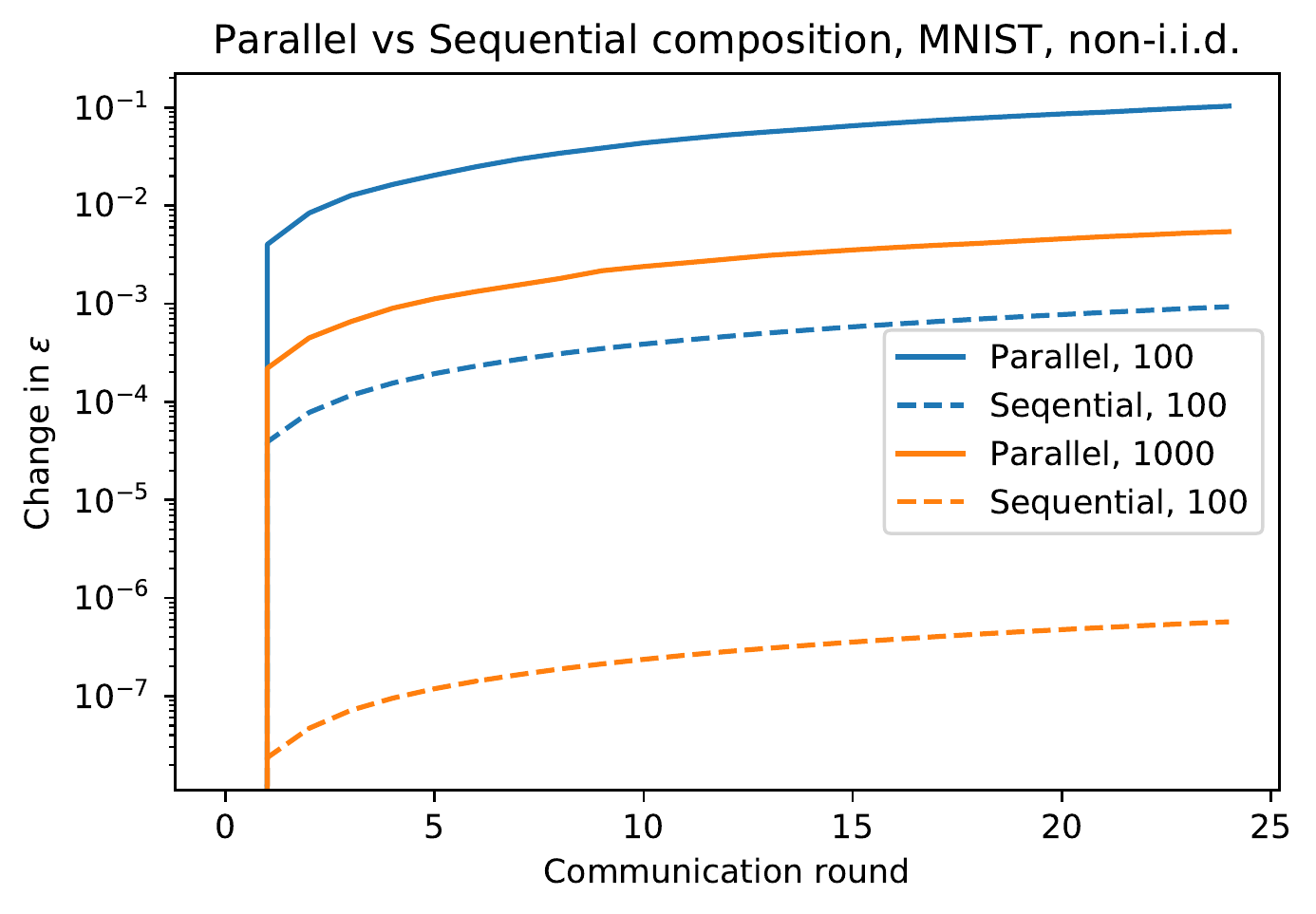}
		\caption{MNIST, non-i.i.d.}
		\label{fig:eps_par_seq_mnist_noniid}
	\end{subfigure}
	\caption{Change in $\varepsilon$ relative to its initial value for parallel and sequential composition modes of instance privacy in the settings of 100 and 1000 clients.}
	\label{fig:eps_par_seq}
\end{figure*}

In this experiment, we test adding client privacy the same way it is done in~\cite{mcmahan2017learning} and~\cite{geyer2017differentially}. We use Bayesian accounting, as described in Section~\ref{sec:client}, and compare it to the classic differential privacy accounting by the moments accountant~\cite{abadi2016deep}. We fix the noise level $\sigma$ and account DP and BDP in parallel.

Tables~\ref{tab:mnist_noniid},~\ref{tab:mnist_iid},~\ref{tab:retina_iid} summarise accuracy and privacy guarantees obtained in this setting for MNIST (non-i.i.d. and i.i.d.) and APTOS respectively. The first column indicates the number of clients, the second---the baseline accuracy of a non-private federated classifier (models described in the previous section). The following columns contain accuracy and privacy parameters obtained for private models using the classic DP and BDP. Despite being trained in parallel, the two techniques may differ in accuracy because in some cases we do early stopping for DP to prevent exceeding privacy budget.

\begin{table}
	\caption{Accuracy and privacy guarantees (reported as a pair $(\varepsilon, \delta)$) on APTOS 2019, i.i.d. setting.}
	\label{tab:retina_iid}
	\centering
	\begin{tabular}{ l | c | c c | c c }
		\toprule
								&  \multicolumn{3}{c|}{Accuracy} 	& \multicolumn{2}{c}{Privacy} 		\\
		\midrule
		{\bf Clients}		& {\bf Baseline}		& {\bf DP} 		& {\bf BDP} 				& {\bf DP} 				& {\bf BDP} 			\\
		\midrule
		100					& $70\%$ 				& $60\%$ 		& $\bm{65\%}$ 		& $(8, 10^{-3})$ 		& $\bm{(2.1, 10^{-3})}$ 	\\
		1K						& $71\%$ 				& $67\%$ 	 		& $\bm{68\%}$ 		& $(2, 10^{-5})$		& $\bm{(0.5, 10^{-5})}$ 	\\
		10K					& $72\%$ 				& $68\%$ 		& $\bm{69\%}$			& $(1, 10^{-6})$ 		& $\bm{(0.2, 10^{-6})}$ 	\\
		\bottomrule
	\end{tabular}
\end{table}

In all cases and for all datasets, we observe substantial benefits of using Bayesian accounting. The accuracy gains are most notable in the non-i.i.d. setting of MNIST, where our method can achieve $10\%$ higher accuracy in the 100 clients setting, because it presents a more difficult learning scenario as explained in the previous section. The privacy gains are consistently significant across all datasets and settings, and taking into account the fact that $\varepsilon$ is exponentiated to get the bound on outcome probability ratios, BDP can reach $e^8/e^2 \approx 400$ times stronger guarantee. Nevertheless, in the settings with few clients, even Bayesian differential privacy does not reach a more comfortable guarantee of $\varepsilon=1$, suggesting that a better privacy-accuracy trade-off may not be feasible due to higher clients identifiability, or that more work is needed in improving training with noise and developing novel privacy mechanisms for federated learning.

Importantly, there is no computation or communication overhead from the users' point of view in these experiments since the privacy accounting code is executed on the server.

\subsection{Instance privacy}
\label{sec:evaluation_instance}

\begin{table}
	\setlength{\tabcolsep}{5pt}
	\caption{Accuracy and privacy guarantees (a pair $(\varepsilon, \delta)$), at instance and client levels, using joint privacy accounting in the setting of 100 clients.}
	\label{tab:joint}
	\centering
	\begin{tabular}{ l | c | c c | c c }
		\toprule
								&  \multicolumn{3}{c|}{Accuracy} 									& \multicolumn{2}{c}{Privacy} 		\\
		\midrule
		{\bf Dataset}		& {\bf Baseline}		& {\bf DP} 	& {\bf BDP} 			& {\bf Client} 								& {\bf Instance} 				\\
		\midrule
		APTOS 2019			& $70\%$ 			& $42\%$ 		& $\bm{64\%}$ 	& $(1, 10^{-3})$ 		& \multirow{3}{*}{$(0.1, 10^{-5})$} 	\\
		MNIST (iid)			& $97\%$ 			& $15\%$ 	 		& $\bm{74\%}$ 	& $(2, 10^{-3})$		& 	\\
		MNIST (non-iid)	& $97\%$ 			& $12\%$ 			& $\bm{62\%}$		& $(4, 10^{-3})$		& 	\\
		\bottomrule
	\end{tabular}
\end{table}

As noted in Section~\ref{sec:instance}, instance privacy is very important in scenarios like collecting medical data from a number of hospitals where patient privacy is at least as crucial as hospital privacy. In this section, we compare two accounting methods proposed earlier: \emph{sequential} and \emph{parallel} accounting.

Depicted in Figure~\ref{fig:eps_par_seq} are the curves showing the growth of $\varepsilon$ estimate with communication rounds. We subtracted the initial value and applied logarithmic scale in order to better show the difference in the rate of growth. Across all settings, it can be seen that parallel accounting leads to faster growth rates, despite the fact that the parallel composition is more efficient (taking maximum over clients instead of a sum). This behaviour can be explained by the fact that each client is unaware of the total dataset size and, having a small number of data points, is convinced that every data example has significant influence on the outcome. The unawareness about other clients in the case of parallel accounting can also explain the fact that we don't observe any improvement in the $\varepsilon$ growth rate with increasing the number of clients. The only exception is the non-i.i.d. MNIST experiment, where the difference is likely to come from increased stability of training and decreased gradient variability with more clients.

The main takeaway from this experiment is that it is beneficial to use sequential accounting for privacy of the federated model whenever communicating the total size of the dataset to users is acceptable. In other cases, and for personal privacy accounting in case of the untrusted curator, parallel accounting can be used, but more noise is necessary for reasonable privacy guarantees.

\begin{figure*}
	\centering
	\begin{subfigure}{0.32\textwidth}
		\includegraphics[width=\linewidth]{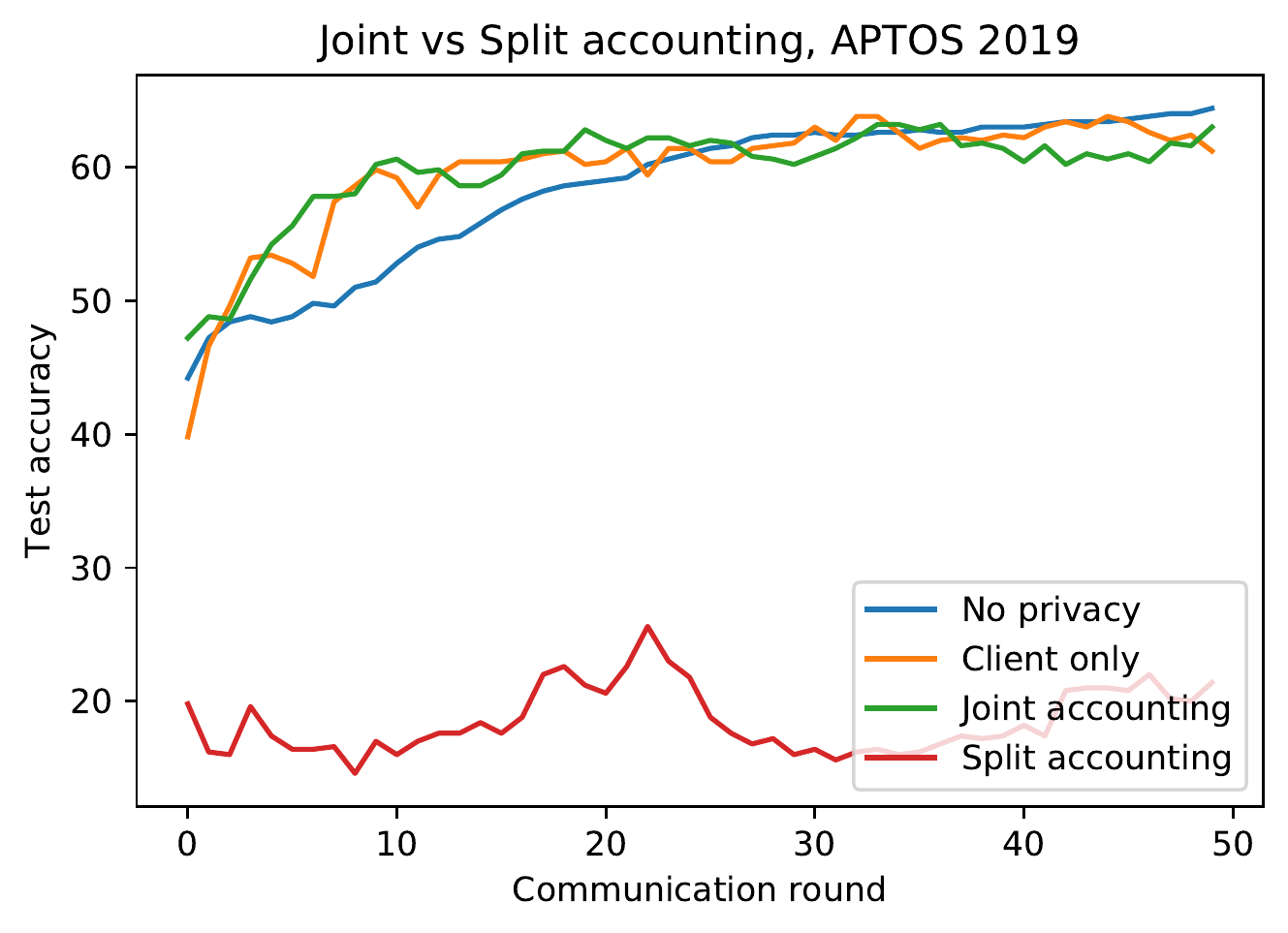}
		\caption{APTOS 2019, 100 clients.}
		\label{fig:recounting_retina}
	\end{subfigure}
	\begin{subfigure}{0.32\textwidth}
		\includegraphics[width=\linewidth]{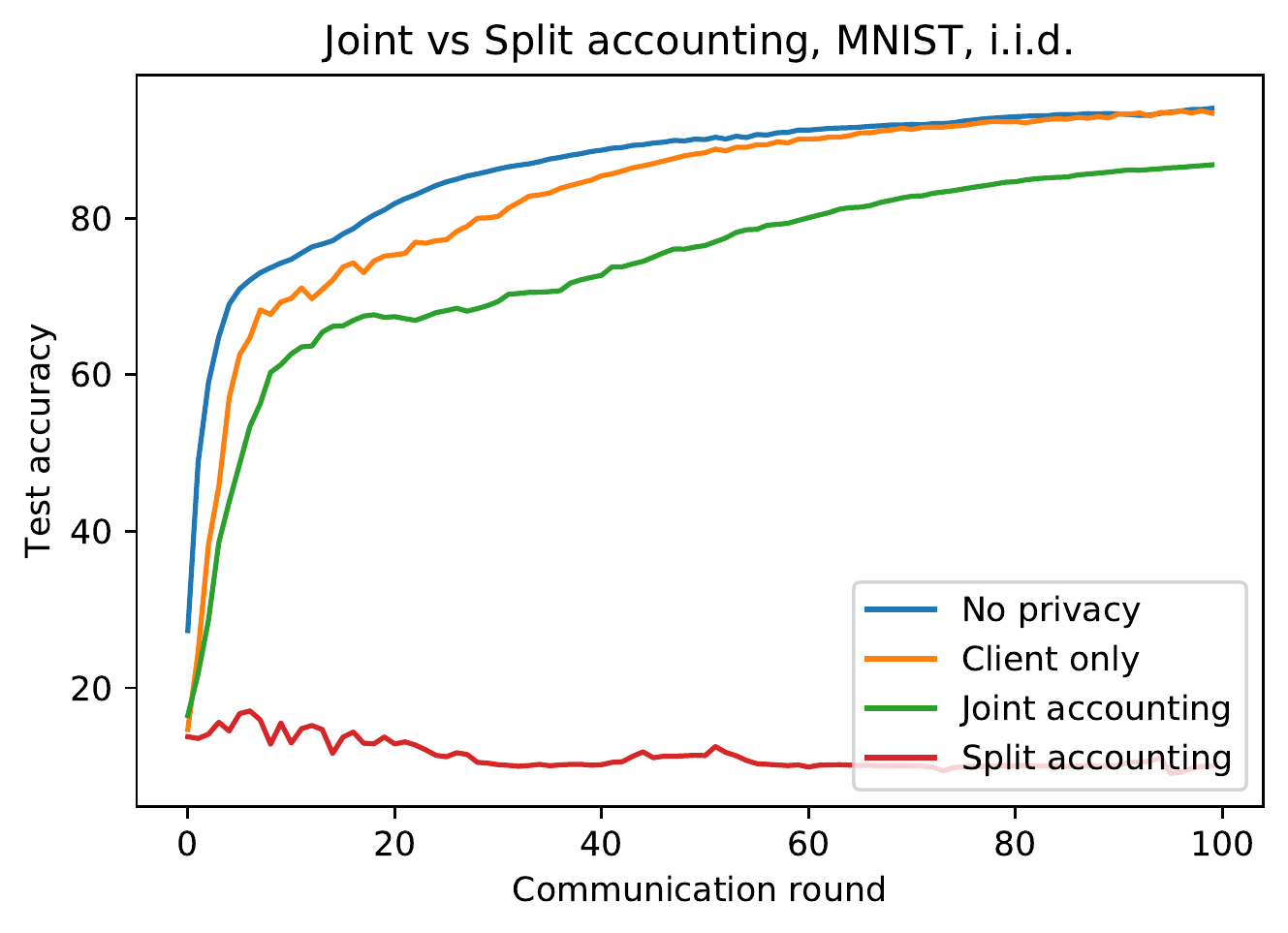}
		\caption{MNIST, i.i.d., 1000 clients.}
		\label{fig:recounting_mnist_iid}
	\end{subfigure}
	\begin{subfigure}{0.32\textwidth}
		\includegraphics[width=\linewidth]{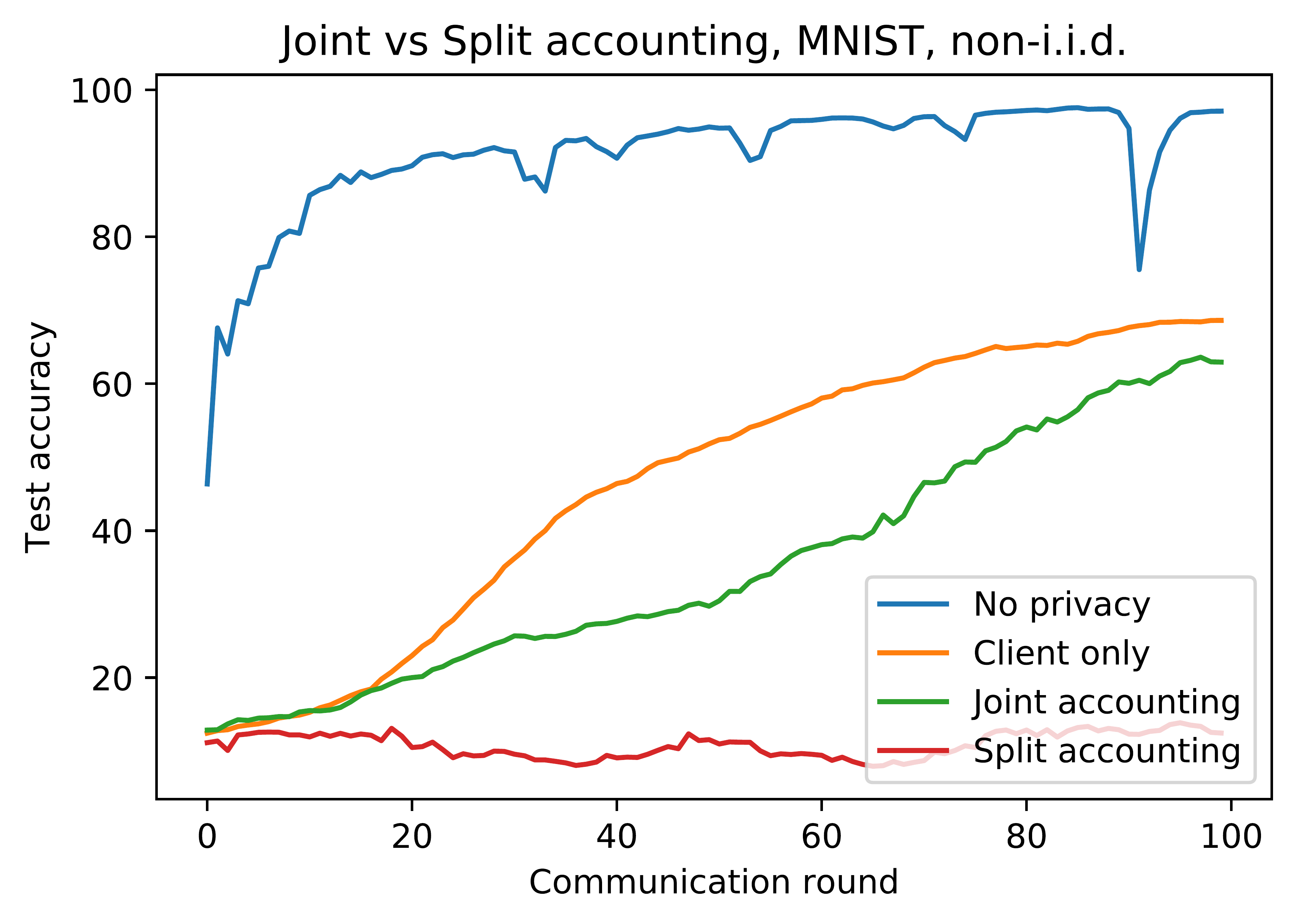}
		\caption{MNIST, non-i.i.d., 1000 clients.}
		\label{fig:recounting_mnist_noniid}
	\end{subfigure}
	\caption{Test accuracy as a function of a communication round for non-private, client-level-only private, and jointly private (using either joint or separate accounting) scenarios.}
	\label{fig:recounting}
\end{figure*}

\subsection{Joint privacy}
\label{sec:evaluation_joint}

Lastly, we would like to test the proposed method of the joint accounting of instance-level and client-level privacy and contrast it to accounting at these two levels separately. We perform experiments in the same settings as above, fixing the client privacy at a certain level ($\varepsilon=1$) and evaluating the speed and quality of training. We also compare to what can be achieved by introducing privacy only at the client level.

Figure~\ref{fig:recounting} displays the test accuracy evolution over communication rounds in the setup of 100 clients for APTOS and 1000 clients for MNIST. The graphs contain curves for training without privacy, client-level-only privacy, and the two accounting paradigms: \emph{joint} and \emph{split}. As expected, the non-private training achieves the best accuracy. Nevertheless, in the i.i.d. setting, client-only private training quickly approaches non-private training in quality. Notably, training with both instance and client privacy using the joint accounting performs nearly as well, while training with the separate accounting completely fails due to excessive amounts of noise at both instance and client levels. For the non-i.i.d. setting, private training is slower, but there is little difference between introducing privacy only at the client level and using the joint accounting at both levels: after a slightly larger number of rounds, training with the joint accounting reaches similar performance. Based on these experiments, we conclude that by using joint accounting we can introduce instance privacy on clients and get client-level privacy at almost no cost.

Finally, we evaluate our method in the strong privacy setting. We set the instance-level privacy to $\varepsilon=0.1$ and stop training when the client privacy reaches the level similar to previous experiments (except APTOS dataset, where we were able to achieve comparable results with lower privacy cost), and report the accuracy that can be achieved in this strict setting. We have also chosen the most difficult scenario of 100 clients. As seen in Table~\ref{tab:joint}, the algorithm with differential privacy performs very poorly on APTOS dataset, and fails to learn on MNIST, in both i.i.d. and non-i.i.d. setting. Its performance is especially affected by the strict instance privacy requirement, since such low levels of $\varepsilon$ necessitate large quantities of noise to be added. It is worth noting, that it might be possible to achieve better results with DP by performing per-example gradient clipping, as in~\cite{abadi2016deep}, but we do not use this technique due to its impracticality. 

On the other hand, our approach manages to achieve reasonable accuracy even under such a strong privacy guarantee. On APTOS dataset, it is just $6\%$ lower than the non-private baseline, while on MNIST, it correctly classifies more than $70\%$ of the test data in the i.i.d. setting and over $60\%$ in the non-i.i.d. setting. One could potentially add more noise on the server and combine the accounting with the instance level noise to slow down the growth of $\varepsilon$ and reach even better performance, but we leave these experiments for future work.

Both instance and joint privacy accounting add some computation overhead on user devices due to multiple gradient calculations. However, performing FL routines when devices are idle and charging, as suggested in~\cite{bonawitz2019towards}, alleviates this problem. Communication overhead is negligible because only a single floating point number is added to user messages.

\section{Conclusion}
\label{sec:conclusion}
We employ the notion of $(\varepsilon, \delta)$-Bayesian differential privacy, a relaxation of $(\varepsilon, \delta)$-differential privacy, to obtain tighter privacy guarantees for clients in the federated learning settings. The main idea of this approach is to utilise the fact that users come from a certain population with similarly distributed data, and therefore, their updates will likely be in agreement with each other. This is a meaningful assumption in many machine learning scenarios because they target a specific type of data (e.g. medical images, emails, motion sensor data, etc.). For example, it may be unjustified to try hiding an absence of an audio record in a training set for the ECG analysis, since the probability of it appearing is in fact much smaller than $\delta$.

We adapt an efficient and tight privacy accounting method for Bayesian differential privacy to the federated setting in order to estimate client privacy guarantees. Moreover, we emphasise the importance of instance-level privacy and propose two variants of privacy accounting at this level. Finally, we introduce a novel technique of joint accounting suitable for obtaining privacy guarantees at instance and client levels jointly from only instance-level noise.

Our evaluation provides evidence that Bayesian differential privacy is more appropriate for federated learning. First, it requires significantly less noise to reach the same privacy guarantees, allowing models to train in fewer communication rounds. Second, the bounds on privacy budget $\varepsilon$ are much tighter, and thus, more meaningful. When the number of clients reaches an order of thousands, which is realistic in many federated learning scenarios, $\varepsilon$ can be kept below $1$. Finally, we demonstrate that by using joint accounting we can get client privacy for free when adding instance privacy. This way, the privacy budget can be kept close to $\varepsilon=1$ for client privacy and $\varepsilon=0.1$ for instance privacy while maintaining reasonably high accuracy.

An important future direction of research is automatically detecting and mitigating scenarios in which tighter privacy guarantees are inapplicable, such as non-stationary data distributions or datasets with non-exchangeable samples.

\bibliographystyle{IEEEtran}
\bibliography{IEEEabrv,bigdata2019}

\end{document}